\pdfoutput=1
\documentclass[11pt]{article}
\usepackage{xcolor}  
\usepackage[preprint]{acl}

\usepackage{times}
\usepackage{latexsym}

\usepackage{tcolorbox}
\tcbuselibrary{breakable}

\usepackage[T1]{fontenc}
\usepackage[utf8]{inputenc}

\usepackage{microtype}

\usepackage{inconsolata}

\usepackage{graphicx}
\usepackage{multirow}
\usepackage{booktabs}
\usepackage{listings} 
\usepackage{xcolor} 
\definecolor{mydarkgreen}{RGB}{0, 150, 0} 
\definecolor{myorange}{RGB}{255, 165, 0}   
\usepackage{float}
\usepackage{asymptote}
\title{Beyond the First Error: Process Reward Models for Reflective Mathematical Reasoning}

 \author{\textbf{Zhaohui Yang}$^{1,2}$\thanks{Equal contribution.},  \textbf{Chenghua He}$^{1*}$, \textbf{Xiaowen Shi}$^3$, \textbf{Linjing Li}$^1$ \\ \textbf{Qiyue Yin}$^1$, \textbf{Shihong Deng}$^2$ \thanks{Corresponding author. Work is done at StepFun.}, \textbf{Daxin Jiang}$^{2}$ \\[7pt]
         $^1$ Institute of Automation, Chinese Academy of Sciences \\ 
         $^2$StepFun, China, $^3$Meituan  \\
         [3pt]
         }
\begin{document}
\maketitle
\begin{abstract}
Many studies focus on data annotation techniques for training effective PRMs. However, current methods encounter a significant issue when applied to long CoT reasoning processes: they tend to focus solely on the first incorrect step and all preceding steps, assuming that all subsequent steps are incorrect. These methods overlook the unique self-correction and reflection mechanisms inherent in long CoT, where correct reasoning steps may still occur after initial reasoning mistakes. To address this issue, we propose a novel data annotation method for PRMs specifically designed to score the long CoT reasoning process. Given that under the reflection pattern, correct and incorrect steps often alternate, we introduce the concepts of \textbf{Error Propagation} and \textbf{Error Cessation}, enhancing  PRMs' ability to identify both effective self-correction behaviors and reasoning based on erroneous steps. Leveraging an LLM-based judger for annotation, we collect 1.7 million data samples to train a 7B PRM and evaluate it at both solution and step levels. Experimental results demonstrate that compared to existing open-source PRMs and PRMs trained on open-source datasets, our PRM achieves superior performance across various metrics, including search guidance, BoN, and F1 scores. Compared to widely used MC-based annotation methods, our annotation approach not only achieves higher data efficiency but also delivers superior performance. Detailed analysis is also conducted to demonstrate the stability and generalizability of our method.

\end{abstract}
\section{Introduction}
Mathematical reasoning has become a crucial metric for evaluating the intelligence of LLMs~\cite{achiam2023gpt,team2024gemini,touvron2023llama}, garnering substantial attention from researchers in recent years. While numerous studies have focused on enhancing LLMs' mathematical reasoning capabilities~\cite{wei2022chain,liu2024augmenting}, LLMs continue to exhibit limitations in practical applications, including calculation errors, flawed derivations, and logical errors. 
PRMs help address these challenges by providing fine-grained evaluation signals for the intermediate steps in the LLM reasoning process, indicating the correctness of each step.

In the short chain-of-thought (CoT) pattern, LLMs lack self-reflection abilities. This means that once a mistake is made, all following steps are likely to be wrong. The PRM only needed to identify correct-to-correct and correct-to-incorrect transitions. Therefore, conventional PRM data construction methods ~\cite{PRM800k,omega-prm} use all steps from correct solutions, but only use steps up to the first error step in  incorrect solutions. However, as shown in Appendix \ref{sec:case}, long CoT LLMs often make mistakes during reasoning, but they can later correct themselves or come up with alternative solutions. Therefore, capturing this incorrect-to-correct pattern is crucial for PRMs to better evaluate the intermediate steps in long CoT reasoning.

Currently, the Monte Carlo (MC) method ~\cite{mathshepherd} is widely used for automatically labeling intermediate steps. However, when dealing with long CoT patterns, the MC method has the following drawbacks: (1) Inaccurate labeling of intermediate reasoning steps. MC method rolls out multiple paths from an intermediate step. If one path reaches the correct answer, it assumes the step is correct. However, long CoT reasoning processes often involve self-correction or reflection, so the final answer doesn’t always reflect the correctness of intermediate steps. (2) High computational cost. The MC method requires multiple rollouts for each intermediate step. In the long CoT pattern, numerous intermediate steps and lengthy rollouts can greatly increase the computational burden.

To address the shortcomings of current PRM designs and MC methods in long CoT scenarios, we propose a new data annotation approach. First, we introduce two fundamental rules specifically for long CoT: \textbf{Error Propagation} and \textbf{Error Cessation}. These rules capture two key patterns: reasoning based on incorrect steps and self-correction after making mistakes. We then incorporate these rules into the LLM judge's prompt to guide the model in annotating intermediate steps of the reasoning process. Finally, we use the annotated data to train the PRM.

Experimental results demonstrate that compared to existing open-source PRMs and PRMs trained on open-source datasets, our PRM achieves the best performance at both the solution level and the step level. At the solution level, we use the traditional Best-of-N (BoN) method to see how well PRMs can select the best answer from several options. Since BoN doesn’t make full use of the process reward signals that PRMs provide, we also propose a new metric that measures whether incorporating these signals into step-level search increases the probability of discovering the correct solution. At the step level, we establish a test set through the cross-validation of two distinct annotation methods (o1 model~\cite{openai2024learning} and manual annotation), enabling rigorous assessment of PRMs in evaluating the correctness of individual solution steps. Across both solution-level and step-level evaluation metrics on the MATH500 and AIME24, our PRM consistently outperforms all baselines. Furthermore, we conduct a comprehensive comparison with the MC-based data annotation method, demonstrating that our method not only achieves greater data efficiency but also delivers superior performance. An in-depth analysis of our method is also conducted to further demonstrate its stability and generalizability.

In summary, our contributions are as follows:
\begin{itemize}
\setlength{\itemsep}{1mm}
\item To the best of our knowledge, we are the first to present a PRM data construction method for long CoT reasoning. We introduce context-aware \textbf{Error Propagation} and \textbf{Error Cessation} mechanisms to effectively capture both wrong-to-wrong and wrong-to-right patterns.

\item We evaluate our method using both solution-level and step-level metrics, demonstrating its effectiveness in evaluating overall solutions as well as individual reasoning steps.
\item Additional analysis demonstrates that our method has notable advantages in robustness, data efficiency, and generalization ability.
\end{itemize}

\section{Related Work}
\paragraph{Long Chain-of-Thought Reasoning Language Models} LLMs have demonstrated remarkable reasoning capabilities for complex tasks. One pivotal method to improve the reasoning ability of LLMs is Chain-of-Thought (CoT) \cite{wei2022chain}, which significantly improves performance by guiding LLMs to generate intermediate reasoning steps. Initial research \cite{wei2022chain, kojima2022large, zhang2022automatic} on CoT mainly focuses on developing effective prompt engineering techniques. Openai o1 \cite{openai2024learning} is the first to introduce inference time scaling law, which employs reinforcement learning to encourage models to generate additional reasoning tokens, thereby overcoming more challenging tasks. In long CoT paradigm, LLMs can decompose problems, explore multiple pathways, and automatically correct reasoning errors. Several efforts \cite{qwq,deepseek-r1,kimi_k1p5, muennighoff2025s1} have successfully replicated this powerful reasoning ability. However, PRMs specifically tailored for this long CoT paradigm remain underexplored.

\paragraph{Application of Reward Models in Mathematical Reasoning}  
Mathematical reasoning in LLMs has seen significant progress with the introduction of reward models. Two types of reward models are commonly used: Outcome Reward Model (ORM) and Process Reward Model (PRM). ORMs assess entire solutions by assigning scores to final answers, while PRMs assign scores to individual steps, offering granular feedback. Research \cite{PRM800k,mathshepherd} demonstrates that PRMs generally outperform ORMs, underscoring their greater potential to enhance reasoning accuracy through guided search \cite{park2024ensembling, zhang2406rest, snell2024scaling} and reinforcement learning \cite{gao2024designing, setlur2024rewarding}. However, the effectiveness of PRMs depends heavily on the availability of high-quality training data, which traditionally requires costly human annotation \cite{PRM800k, uesato2022solving}. To address this challenge, recent work \cite{mathshepherd, omega-prm, mips, chen2024alphamath, PRMlessons} explores automated techniques for collecting data, streamline the data collection process based on MC estimation. Despite these advances, substantial computational demands of MC-based methods remain a significant barrier when dealing with long reasoning chains, hindering the accumulation of sufficient training data.

\section{Disadvantages of Current PRMs}

Previous works on training PRMs primarily focus on the first incorrect step and assume that all subsequent steps are reasoned based on this erroneous step, thus deeming them all incorrect. However, in a long CoT pattern, self-correction behaviors often occur, where correct reflective steps can still follow an incorrect step. This results in a distribution shift between the PRM training data and the data encountered during inference in long CoT scenarios,  potentially affecting performance. Therefore, we aim to investigate:  \textit{Does the self-correction behavior of long CoT models impact the performance of current PRMs?}

To answer this question, we construct two test sets: an Error-Free Set (EF Set) and a Reflection-Based Set (RB set). The final answers of solutions for both sets are correct. The key difference lies in the correctness of the intermediate reasoning steps. All reasoning steps in the Error-Free Set are correct, while solutions in the Reflection-Based Set contain erroneous intermediate steps and perform self-correction. We evaluate several open-source PRMs on these two test sets: Qwen2.5-Math-PRM-7B \cite{PRMlessons}, MathShepherd-PRM-7B \cite{mathshepherd}, and Skywork-PRM-7B \cite{skyworkopeno12024}, focusing on the accuracy of PRMs in determining whether the  solutions are correct.

Our experimental results, as presented in Table \ref{table:motivation_exp1}, reveal significant performance disparities between the two test sets across all evaluated PRMs. The most effective model, Qwen2.5-PRM-7B, demonstrates a 10\% performance gap between two test sets, while other models show even larger disparities, even exceeding 50\%. These findings suggest that PRMs exhibit reduced effectiveness in evaluating reasoning processes that incorporate intermediate reflection.

\begin{table}[htbp]
\centering
\begin{tabular}{lcc}
\toprule
\multicolumn{1}{c}{\textbf{Model}} & \textbf{EF Set} & \textbf{RB Set}  \\
\hline
Qwen2.5-MATH-PRM-7B  & 0.98 & 0.88 \\
MathShepherd-PRM-7B & 0.79 & 0.27\\
Skywork-PRM-7B & 0.93 & 0.59\\
\bottomrule
\end{tabular}
\caption{The prediction accuracy of open-source PRMs.}
\label{table:motivation_exp1}
\end{table}

\section{Method}

\begin{figure*}[!htbp]
    \centering
    \includegraphics[width=\textwidth]{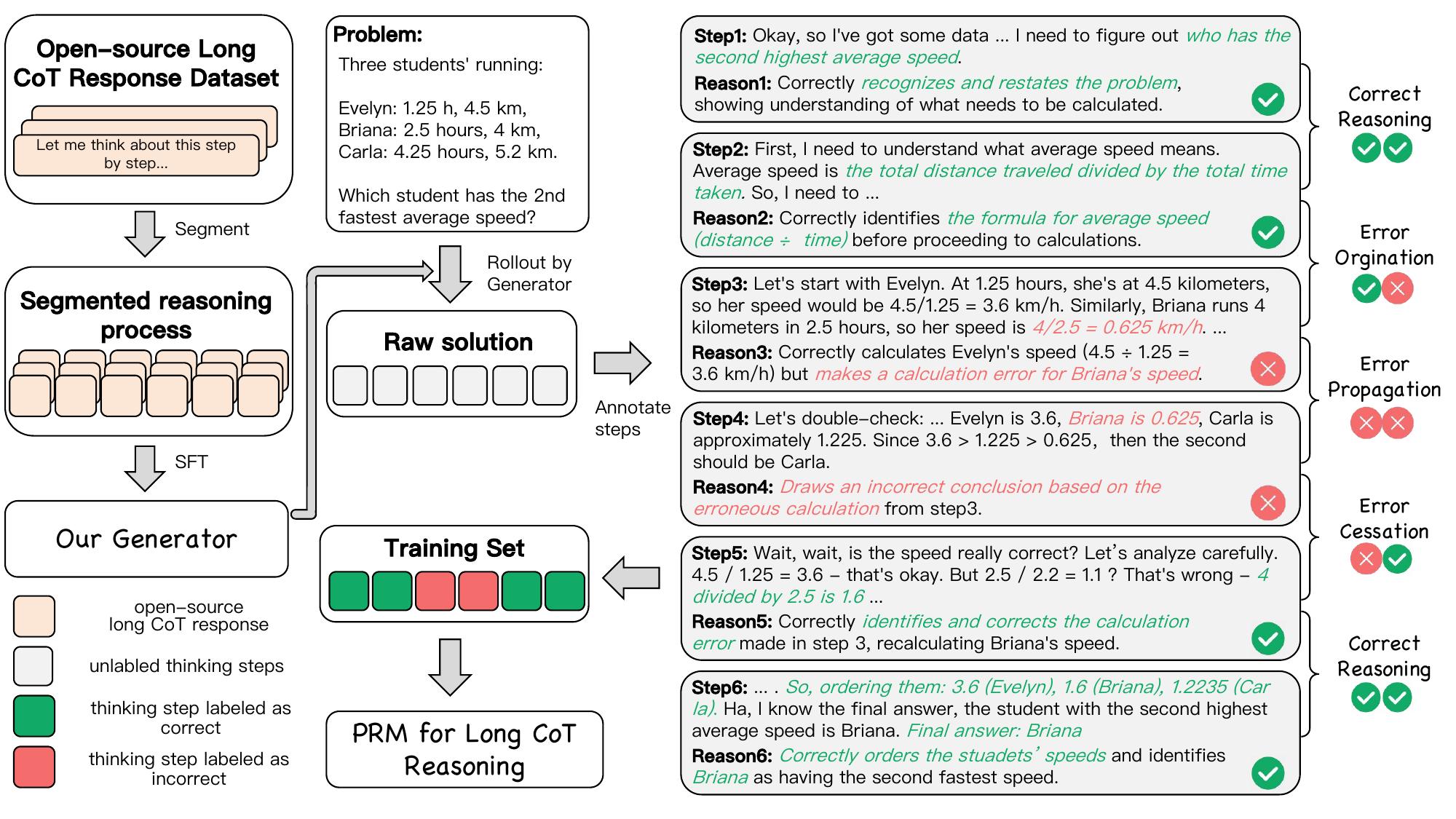}
    \caption{The overall framework of our method.}
    \label{fig:paradigm}
\end{figure*}

In this section, we introduce a framework for constructing PRM data tailored to reflective reasoning. First, we introduce how to train a long CoT model that can autonomously segment its output based on semantics. Then, we describe our detailed step-by-step annotation guidelines and the implementation of PRM training. The overall framework of our method is shown in Figure \ref{fig:paradigm}.

\subsection{Dividing Reasoning Process into Steps}
\label{sec:division_solutions}
Currently, no open-source LLMs satisfies both requirements: (1) generating long CoT reasoning processes, and (2) including separators ensuring semantic integrity at each step. While using double line breaks as delimiters to segment reasoning chains is a common practice, it compromises semantic coherence. Moreover, in the long CoT paradigm, this approach results in numerous segmentation steps, increasing annotation workload.

To address this challenge, we develop a specialized generator based on Supervised Fine-Tuning (SFT). For constructing the SFT dataset, inspired by \cite{zheng2024processbench}, we adopt a two-step approach to segment open-source long CoT data. We first replace all line breaks with spaces, and then utilize LLMs to resegment the reasoning process. Our segmentation balances cognitive cohesion (merging conceptually related steps) and modular independence (ensuring each step represents a distinct reasoning unit), while controlling step count and token length. A Detailed solution segmentation case is shown in Appendix \ref{sec:solution_reformatting}.

\subsection{Annotation Standards for Reflective Reasoning}
\label{sec:annotation_standards}

In the long CoT reasoning process, it is common for LLMs to revisit previous steps for self-correction and reflection, which is important for improving reasoning ability. However, previous annotation methods typically focus only on identifying the first erroneous step and assessing the correctness of preceding steps. These approaches overlook effective reflections after the first error in reflective reasoning, which hampers comprehensive process supervision. 

We classify steps following an error into two types: (1) Reasoning on faulty assumptions, which compounds mistakes and leads further from the solution; (2) Corrective steps, which identify and address errors, guiding the reasoning back on track.
Based on our categorization, we introduce two new annotation rules designed to capture these post-error reasoning dynamics:
\begin{itemize}
    \item \textbf{Error Propagation}: If the previous steps are incorrect and the current step neither introduces a new approach nor corrects the previous mistakes, but instead builds upon the erroneous steps, the current step is also considered incorrect.
    \item \textbf{Error Cessation}: If the previous steps are incorrect but the current step introduces a new, error-free approach or corrects the previous mistakes, the current step is considered correct.
\end{itemize}

Based on the annotation rules above, we assign appropriate labels to each step in the reasoning process. In addition, while our primary focus is on mathematical reasoning problems, these rules can also be applied to other domains such as coding, the 24-point game, and more. 

\subsection{LLM Judgement}
\label{sec:llm_judgement}
Reflective models like the o1 \cite{openai2024learning} series have demonstrated outstanding performance in complex reasoning tasks. The results of the manual inspection in Appendix \ref{sec:Human Annotation Details} indicate that reasoning LLMs are capable of effectively solving annotation tasks. Therefore, we incorporate the rules in Section \ref{sec:annotation_standards} into the prompt (see Appendix \ref{sec:Prompt Template for LLM Judger}), allowing the reflective LLM to evaluate the correctness of each step. 

\subsection{Process Reward Model}
Since the label for each step is a binary score, we use standard classification loss to train our PRM:
\begin{equation}
    L_{PRM}=\sum_{i=0}^{K}  \hat{y}_{i}\mathrm{log} {y_i}+(1-\hat{y}_i)\mathrm{log}(1-y_{i})
\end{equation}
where $y_i$ represents the golden label of the i-th step $s_i$, $\hat{y}_i=PRM(\mathrm{prompt},s_{<=i})$ is the predicted score for $s_i$ by PRM, and $K$ is the total number of steps of the solution.
\section{Experiments}
In this section, we present the experimental results of our PRM in comparison with (1) open-source PRMs and PRMs trained on open-source datasets (Section\ref{sec:main_res1}), and (2) PRMs trained on data generated through MC-based methods (Section\ref{sec:main_res2}).

\subsection{Experiment Settings}
\begin{table*}[ht]
\centering
\resizebox{\linewidth}{!}{
\begin{tabular}{@{}lccccccc@{}}
\toprule
\multicolumn{1}{@{}c}{\multirow{2}{*}{\textbf{Model}}}  & \multicolumn{2}{c}{\textbf{MATH500}} & \multicolumn{2}{c}{\textbf{AIME2024}} & \multicolumn{3}{c}{\textbf{Step-Level Testset}}\\
\multicolumn{1}{@{}c}{} & PRM@64 & PRM@8-step & PRM@64 & PRM@8-step  & Precision & Recall & F1 \\
 \hline
 PRM-PRM800K & 0.712 & 0.682 & 0.133 &  0.067 & 0.640 & 0.963 & 0.758 \\
 PRM-MS&  0.758 &  0.706& 0.233 &  0.100 &  0.613 & \textbf{0.994} & 0.758 \\
  \hline
Qwen2.5-PRM-7B& 0.776 & 0.738 & 0.167 & 0.133 & 0.634 & 0.972 & 0.768\\
MathShepherd-7B& 0.778 & 0.702 & \textbf{0.267} & 0.100 & 0.863 & 0.376 & 0.523\\
Skywork-PRM-7B& 0.754 & 0.740 & 0.133 & 0.100 &  \textbf{0.936} & 0.351 & 0.512 \\
\hline
 Ours & \textbf{0.816} & \textbf{0.750} & \textbf{0.267} & \textbf{0.167} & 0.850 & 0.806 & \textbf{0.828}\\
\toprule
\end{tabular}
}
\caption{The solution-level and step-level performance of Qwen2.5-7B-SFT$^{*}$ using our PRM and other baselines.}
\label{tab:overall_performance}
\end{table*}

\subsubsection{Dataset}
\label{sec:dataset}
\textbf{Training:}
For training the PRMs and the generator, we construct the prompt set by combining the MATH training set with AIME problems from 1983 to 2023.

\textbf{Evaluation:}
Our test set consists of MATH500 \cite{hendrycks2021measuring} combined with AIME2024 \cite{aime}. 
Additionally, we employ the generator to produce 800 solutions for prompts in the test set, forming our step-level test set, which is used to evaluate the accuracy of PRM in assessing each step of the reasoning process. The labels for this set are determined through a combination of cross-validation with the o1 model and manual annotation. For more information, please refer to Appendix~\ref{sec:Human Annotation Details}.

\subsubsection{Generator}
\label{sec:generation_model}
We segment about 2,000 samples in open-source datasets \cite{min2024imitate} using method described in Section~\ref{sec:division_solutions}, and fine-tuned the Qwen2.5-7B-base \cite{qwen2.5} to obtain Qwen2.5-7B-SFT$^{*}$.

\subsubsection{Metrics}
Our evaluation of PRMs includes solution-level and step-level analyses. At solution level, we assess PRMs' ability to identify high-quality solutions using PRM@N and PRM@N-step. At step level, we examine PRMs' accuracy to identify correctness of individual steps with F1-related metrics.

\noindent \textbf{Best of N.}
Consistent with previous work \cite{PRM800k, mathshepherd, mips, omega-prm}, we use Best-of-N for evaluation, which selects the highest-scoring response from N candidate responses. We define this evaluation metric as PRM@N. The score of each candidate response is determined by the score at the final step.

\noindent \textbf{Online Search.} During each step, N candidates are sampled and the highest-scoring step is chosen to continue the generation. 
We define this metric as PRM@N-step.

\noindent \textbf{Classification Metrics.}
To evaluate the classification performance at the step level, we use precision, recall, and F1 score as our metrics.

\subsubsection{Baselines} 

Our baseline models are categorized as follows:
\begin{itemize}
    \item \textbf{Fine-Tuned PRMs}: PRMs fine-tuned on open-source datasets, including: (1) PRM-PRM800K: Fine-tuned with the PRM800K dataset from OpenAI~\cite{PRM800k}. (2) PRM-MS: Fine-tuned with the Math Shepherd dataset from DeepSeek~\cite{mathshepherd}.
    \item \textbf{Open-Source PRMs}: Existing open-source models, including: (1) Qwen2.5-MATH-PRM-7B (abbreviated as Qwen2.5-PRM-7B)~\cite{PRMlessons}. (2) DeepSeek-MathShepherd-7B (abbreviated as MathShepherd-7B)~\cite{mathshepherd}. (3) Skywork-PRM-7B~\cite{skyworkopeno12024}.
\end{itemize} 

\subsection{Main Results}
\label{sec:main_res1}

\paragraph{Solution Level}
According to the results shown in Table \ref{tab:overall_performance}, we find that: (1) Our PRM consistently outperforms other PRMs in PRM@64. Specifically, on MATH500, it achieves an accuracy of 81.6\%, which is 3.8\% higher than the second-best PRM, demonstrating enhanced capability in identifying high-quality solutions. (2) Our PRM performs the best in PRM@8-step. This indicates that our PRM is capable of providing better intermediate signals which guide the generator to produce higher-quality solutions.

\paragraph{Step Level}
The step-level experimental results in Table \ref{tab:overall_performance} indicate that: (1) Our PRM performs the best in step-level evaluation, achieving the highest F1 score.
(2) Both open-source PRMs and PRMs trained on current open-source datasets exhibit an imbalance between precision and recall. PRM-PRM800K, PRM-MS, and Qwen2.5-PRM-7B tend to classify incorrect steps as correct, while MathShepherd-7B and Skywork-PRM-7B are more prone to classifying correct steps as incorrect. In contrast, our PRM demonstrates the most balanced performance. 

\begin{figure*}[htbp]
  \centering
  \includegraphics[width=1.8\columnwidth]{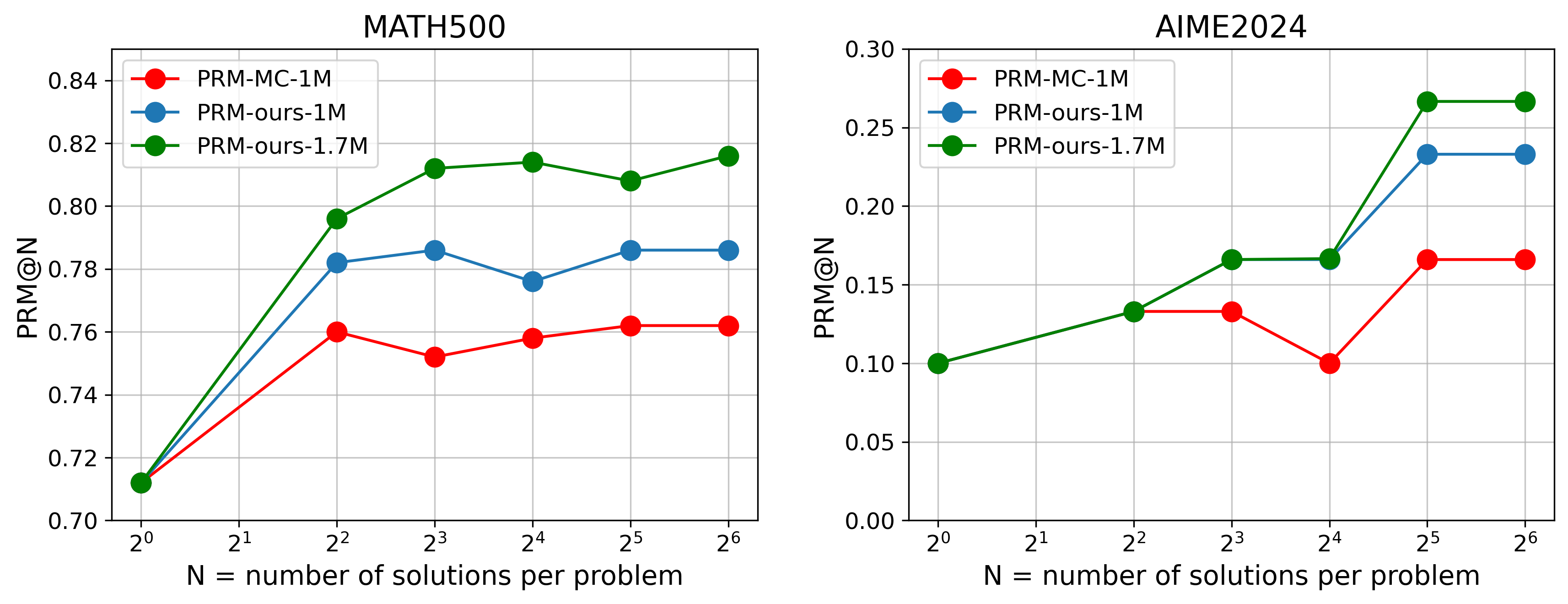}
  \caption{PRM@N of Qwen2.5-7B-SFT$^{*}$ using PRMs trained on data annotated by MC-based and our method.}
  \label{fig:mc_experiment}
\end{figure*}

\subsection{Comparison with MC-based Methods}
\label{sec:main_res2}
Although MC-based method has been widely used, it contains significant noise. Their effectiveness relies on the ability of completion models. Correct steps may be misjudged as incorrect when completion models fail to produce correct solutions within limited rollouts. Reflective reasoning paradigm exacerbates the annotation noise. Incorrect steps may be misjudged as correct due to subsequent reflective and error-correction behaviors \cite{lanham2023measuring}. 

\subsubsection{Experiment Setup}

For each prompt in training set, we select 8 solutions generated by Qwen2.5-7B-SFT$^{*}$ for MC-based annotation, with the total number of steps approaching 1 million. We employ Qwen2.5-7B-SFT$^{*}$ to perform 8 completions for each step to assess their correctness. Following previous work~\cite{mathshepherd}, we create hard labels based on the sampled completions and train PRM-MC-1M.  
The annotation process consumes approximately 23,040 A100 GPU hours. To ensure a fair comparison with PRM-MC-1M, we select the subset of our training data corresponding to solutions used in PRM-MC-1M training to train PRM-ours-1M. The PRM trained on the entire training set is referred to as PRM-ours-1.7M.
\subsubsection{Results}

Figure \ref{fig:mc_experiment} presents the experimental results, which reveal that:
(1) PRM-ours-1M achieves superior PRM@N scores compared to PRM-MC-1M consistently, demonstrating the improved quality of process supervision signals generated by our method. (2) PRM@N for PRM-ours-1.7M consistently outperforms PRM-ours-1M, indicating robust scalability of our approach. The PRMs trained using our method can effectively leverage more process supervision signals in larger training datasets.

Additionally, leveraging LLM for data annotation significantly improves efficiency compared to the MC-based method, enabling faster data annotation and more rapid training iterations.

\section{Analysis}
\subsection{Robustness Compared to MC-based Method}
This section aims to demonstrate that our method can provide more robust process supervision signals compared to commonly used MC-based methods. First, we  show that there is a strong correlation between the signals obtained via the MC-based method and the completion model, and the correlation strengthens with increasing reasoning chain length. Second, we will compare the performance gap when the same PRM provides process reward signals to different models.

\paragraph{Correlation between Signals and Completion Model} 

\begin{figure}[t]
    \centering
    \includegraphics[width=\linewidth]{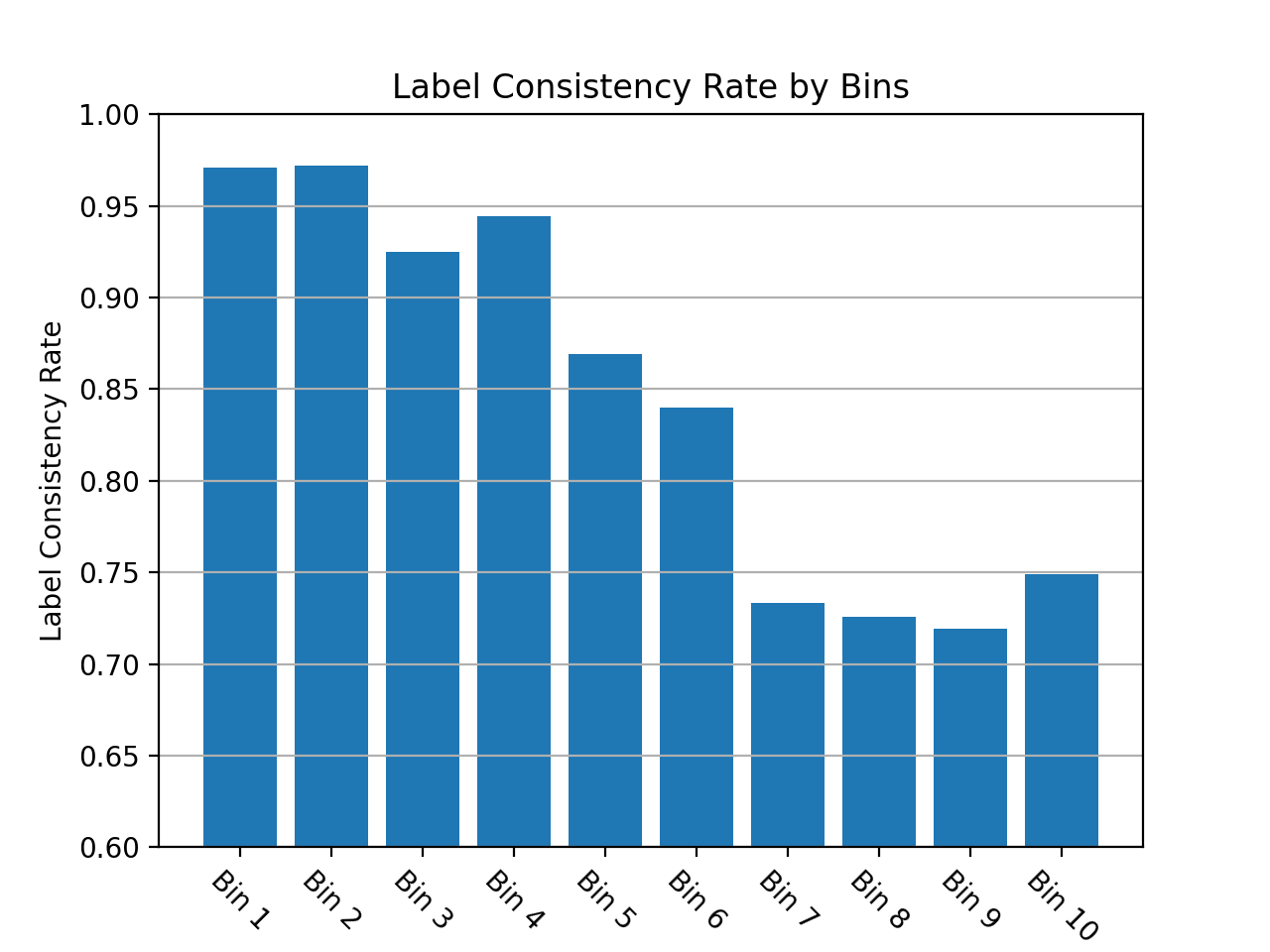}
    \caption{We categorize 1,000 solutions into 10 equal-sized bins based on their step counts, with Bin 1 containing solutions with the fewest steps and Bin 10 containing those with the most steps. Within each bin, we calculate the proportion of steps where both completion models assign identical hard labels.}
    \label{fig:Fig_robustness}
\end{figure}

We first fine-tune LLaMA3.1-8B-Base \cite{dubey2024llama} using the SFT data described in Section \ref{sec:generation_model} to get LLaMA3.1-8B-SFT$^{*}$, both Qwen2.5-7B-SFT$^{*}$ and LLaMA3.1-8B-SFT$^{*}$ are employed as completion models for our analysis. Subsequently, we randomly sample 1,000 solutions generated by Qwen-2.5-7B-SFT$^{*}$ and use both completion models to annotate step-level hard labels following Math-Shepherd. Our analysis reveals that only 79\% of the solution steps receive identical annotations from both models. Moreover, we observe a negative correlation between solution length and inter-model consistency. As illustrated in Figure \ref{fig:Fig_robustness}, the consistency rate between the two completion models decreases as the number of solution steps increases.  This finding suggests that MC-based annotation methods become progressively less reliable as reasoning chains extend.

\paragraph{Performance Gap when Providing Signals for Different Models} 

\begin{table}[t]
\centering
\resizebox{\linewidth}{!}{
\begin{tabular}{@{}lcccccc@{}}
\toprule
\multicolumn{1}{@{}c}{\multirow{2}{*}{\textbf{Metric}}} & \multicolumn{3}{c}{\textbf{PRM-MC-1M}} & \multicolumn{3}{c}{\textbf{PRM-ours-1M}} \\
\multicolumn{1}{@{}c}{} & Qwen & LLaMA & Gap & Qwen & LLaMA & Gap \\
\hline
PRM@8  & 0.752 & 0.386 & 0.366 & 0.786 & 0.466 & 0.320  \\
PRM@16 & 0.758 & 0.370 & 0.388 & 0.776 & 0.464 &  0.312 \\
PRM@32 & 0.762 & 0.400 & 0.362 & 0.786 & 0.478 &  0.308 \\
PRM@64 & 0.760 & 0.436 & 0.324 & 0.786 & 0.518 &  0.268 \\
\toprule
\end{tabular}
}
\caption{PRM@N of different models using PRM-MC and PRM-ours, along with the performance gaps.}
\label{table:performace gap}

\end{table}
We evaluate PRM-MC-1M and PRM-ours-1M by using them to provide reward signals during the inference processes of LLaMA3.1-8B-SFT$^{*}$ and Qwen2.5-7B-SFT$^{*}$ on MATH500. The results in Table~\ref{table:performace gap} show that PRM-MC-1M introduces a significantly larger performance gap between the two models compared to PRM-ours-1M. This suggests that our method produces a more robust PRM, capable of providing consistent reward signals across different language models. Moreover, PRM-ours-1M demonstrates superior PRM@N compared to PRM-MC-1M across both Qwen2.5-7B-SFT$^{*}$ and LLaMA3.1-7B-SFT$^{*}$. This consistent performance advantage across different models validates the adaptability of our PRM on different generators.

\subsection{Data Distribution}

This section examines the distributional differences between our training data and existing open-source datasets. 

\begin{figure*}[htbp]
  \centering
  \includegraphics[width=2.0\columnwidth]{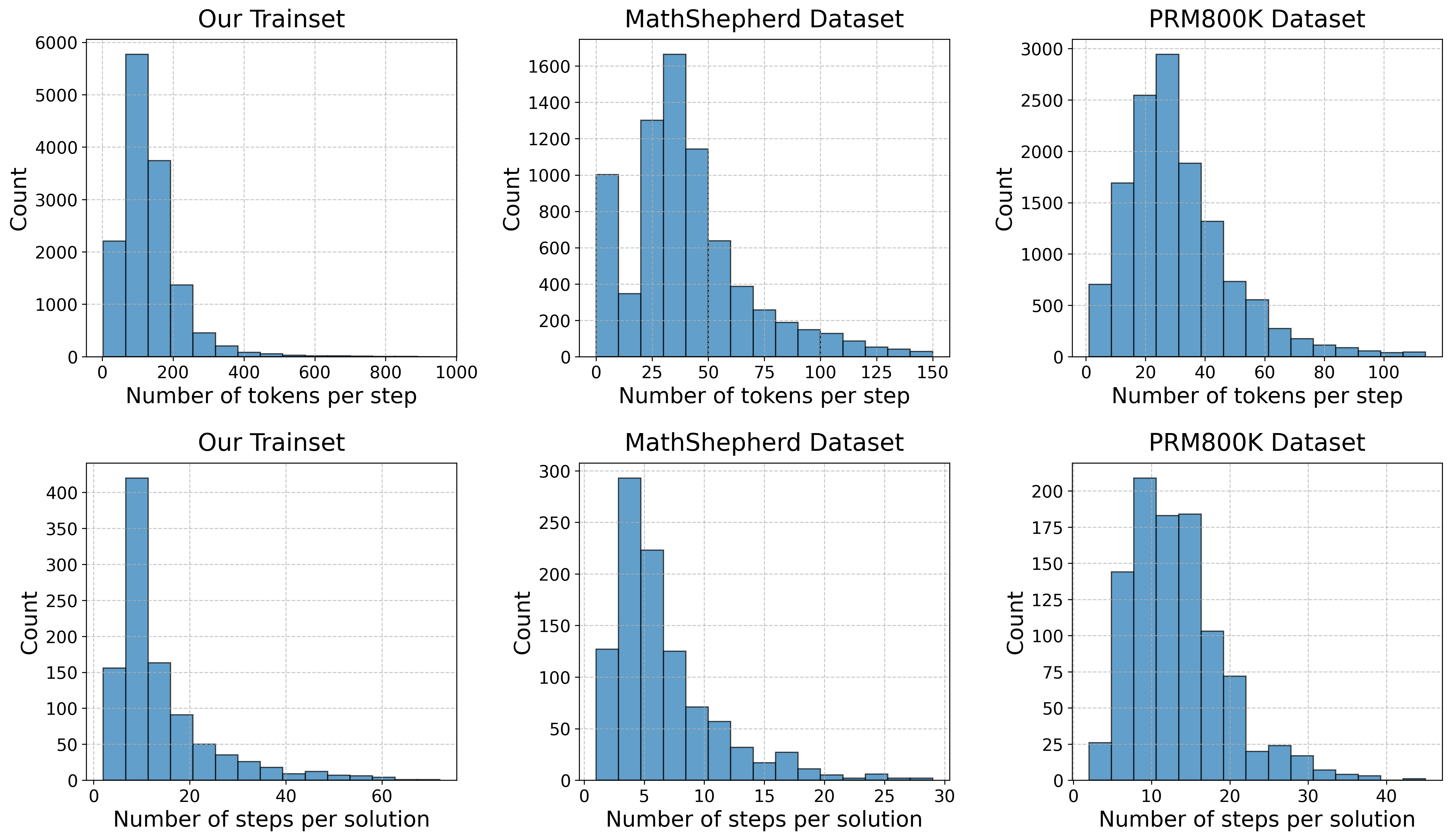}
  \caption{Distribution of the number of steps in each solution and the number of tokens contained in each step across different datasets. We randomly select 1,000 samples from each dataset for statistical analysis.}
  \label{fig:step_and_token_distribtuion}
\end{figure*}

As illustrated in Figure \ref{fig:step_and_token_distribtuion}, our dataset exhibits a significantly higher number of steps per solution and greater token consumption per step compared to open-source datasets.
This distributional divergence primarily stems from our annotated data, which comprises long reasoning chains that incorporate reflective processes. These chains inherently demand more elaborate step-by-step reasoning and extended cognitive operations.

We believe that this distributional discrepancy causes PRMs trained on open-source data to generate less effective process reward signals for reflective reasoning chains. This limitation appears to be one of the key factors contributing to their inferior performance compared to our PRM at both solution level and step level.
\subsection{Enhanced Data Utilization Beyond First-Error Steps}

\begin{table}[t]
\centering
\resizebox{\linewidth}{!}{
\begin{tabular}{clccc}
\toprule
\multicolumn{1}{c}{\textbf{Datasets}} & \multicolumn{1}{c}{\textbf{Metrics}} & \textbf{FE} & \textbf{FES} & \textbf{Ours} \\
\hline
\multirow{4}{*}{MATH500} & PRM@8 & 0.764 & 0.770 & \textbf{0.786} \\
                      & PRM@16 & 0.774 & \textbf{0.788} & 0.776 \\
                      & PRM@32 & 0.776 & \textbf{0.792} & 0.786 \\
                      & PRM@64 & 0.780 & \textbf{0.792} & \textbf{0.792} \\
\hline
\multirow{4}{*}{AIME2024} & PRM@8 & 0.133 & 0.133 & \textbf{0.167} \\
                      & PRM@16 & 0.133 & 0.133 & \textbf{0.167} \\
                      & PRM@32 & 0.167 & 0.167 & \textbf{0.233} \\
                      & PRM@64 & 0.200 & 0.167 & \textbf{0.233} \\
\toprule
\end{tabular}}
\caption{PRM@N of Qwen2.5-7B-SFT$^{*}$ using PRMs trained under different experimental settings.}
\label{table:data utilization compare}
\end{table}

This section aims to demonstrate the advantages of our method over the traditional paradigm, which typically only considers the first error step. Our approach extracts more training signals from each solution while maintaining the same sample size.  We design a comparative study with three experimental settings: (1) Baseline: Our method; (2) First-error truncation (FE): Using identical training samples as (1) but only considering steps until the first error occurs; and (3) First-error truncation with supplementary data (FES): Extending (2) with additional training samples to align the total training signals in (1).

The experimental results in Table~\ref{table:data utilization compare} reveal two key findings. First, given an equal number of solution samples, our method achieves superior performance on both MATH500 and AIME2024 by utilizing additional process supervision signals from each solution. Second, when aligning the number of signals, our method achieves comparable results to traditional paradigm while using fewer solutions. Notably, on the more challenging AIME2024, our method even demonstrates a slight advantage. The enhanced effectiveness on AIME2024 can be attributed to its inherently challenging nature, which demands more sophisticated reflective reasoning processes.

\subsection{Out-of-Distribution Performance}

\begin{table}[t]
\centering
\begin{tabular}{@{}ccc@{}}
\toprule
\multicolumn{1}{c}{\textbf{Model}} & \textbf{GSM8K} & \textbf{OBen}  \\
\hline
PRM-PRM800K       & 0.914 & 0.394  \\
PRM-MS               & \textbf{0.954} & 0.441 \\
\hline
Qwen2.5-PRM-7B       & 0.950 & 0.458 \\
MathShepherd-7B      & \textbf{0.954} & 0.455 \\
Skywork-PRM-7B       & \textbf{0.954} & 0.446 \\
\hline
Ours          & 0.950 & \textbf{0.510} \\
\toprule
\end{tabular}
\caption{PRM@64 on GSM8K and OBen of Qwen2.5-7B-SFT$^{*}$ using different PRMs.}
\label{table:Out-of-Distribution Performance}
\end{table}

To demonstrate the generalizability of our method, we conduct evaluations on both elementary and advanced mathematical problems. Our evaluation select 500 questions from GSM8K~\cite{cobbe2021training} representing simple math problems, and 675 questions from Olympiad Bench (OBen)~\cite{he2024olympiadbench} which contains challenging, competition-level problems.

As can be seen from Table~\ref{table:Out-of-Distribution Performance}, most PRMs demonstrate comparable performance on GSM8K. While our PRM performs marginally below PRMs trained on Math-Shepherd and Skywork-PRM-7B, this slight difference might be attributed to the inclusion of GSM8K examples in their training set. Notably, when evaluated on the more challenging OBen dataset, our method exhibits substantially superior performance compared to all baseline approaches, highlighting its strength in handling sophisticated math problems. More information about the performance advantages of our PRM on complex datasets can be found in Appendix \ref{Appendix H}.

\section{Conclusion}
In this paper, we introduce a PRM data annotation technique designed for reflective reasoning processes. We propose concepts of \textbf{Error Propagation} and \textbf{Error Cessation}, which enable precise identification of steps based on flawed premises and highlight moments of meaningful reflection. Additionally, leveraging LLMs for annotation helps reduce the resource burden caused by extended reasoning chains. Experimental results demonstrate our PRM outperforms current open-source PRMs and PRMs trained on open-source datasets at solution and step levels. Compared to commonly-used MC-based methods, our method also exhibits comprehensive superiority. 

\section{Limitaions}
Despite achieving superior performance in solution-level and step-level metrics compared to other baselines, there are several limitations to our approach. (1) Scalability: While our PRM provides more accurate signals in our diverse evaluation test, due to constraints in training data and computational resources, we are unable to validate the advantages of our method in broader experimental settings. Future work could explore how the accuracy and generalization of PRM scale as the number of prompts and generated solutions increases. (2) Dependency on LLM Capabilities: Our labeling method requires the LLM judges to have strong reasoning capabilities. Consequently, the accuracy of labeling is limited by the inherent abilities of the model itself. However, with ongoing improvements in open-source reflective models \cite{qwq, deepseek-r1,muennighoff2025s1}, this issue could be mitigated. Future work could also investigate the performance of small reflection models used for annotation.

\bibliography{custom}

\begin{thebibliography}{35}
\providecommand{\natexlab}[1]{#1}

\bibitem[{Achiam et~al.(2023)Achiam, Adler, Agarwal, Ahmad, Akkaya, Aleman, Almeida, Altenschmidt, Altman, Anadkat et~al.}]{achiam2023gpt}
Josh Achiam, Steven Adler, Sandhini Agarwal, Lama Ahmad, Ilge Akkaya, Florencia~Leoni Aleman, Diogo Almeida, Janko Altenschmidt, Sam Altman, Shyamal Anadkat, et~al. 2023.
\newblock Gpt-4 technical report.
\newblock \emph{arXiv preprint arXiv:2303.08774}.

\bibitem[{Chen et~al.(2024)Chen, Liao, Li, and Fan}]{chen2024alphamath}
Guoxin Chen, Minpeng Liao, Chengxi Li, and Kai Fan. 2024.
\newblock Alphamath almost zero: process supervision without process.
\newblock \emph{arXiv preprint arXiv:2405.03553}.

\bibitem[{Cobbe et~al.(2021)Cobbe, Kosaraju, Bavarian, Chen, Jun, Kaiser, Plappert, Tworek, Hilton, Nakano et~al.}]{cobbe2021training}
Karl Cobbe, Vineet Kosaraju, Mohammad Bavarian, Mark Chen, Heewoo Jun, Lukasz Kaiser, Matthias Plappert, Jerry Tworek, Jacob Hilton, Reiichiro Nakano, et~al. 2021.
\newblock Training verifiers to solve math word problems.
\newblock \emph{arXiv preprint arXiv:2110.14168}.

\bibitem[{Dubey et~al.(2024)Dubey, Jauhri, Pandey, Kadian, Al-Dahle, Letman, Mathur, Schelten, Yang, Fan et~al.}]{dubey2024llama}
Abhimanyu Dubey, Abhinav Jauhri, Abhinav Pandey, Abhishek Kadian, Ahmad Al-Dahle, Aiesha Letman, Akhil Mathur, Alan Schelten, Amy Yang, Angela Fan, et~al. 2024.
\newblock The llama 3 herd of models.
\newblock \emph{arXiv preprint arXiv:2407.21783}.

\bibitem[{Gao et~al.(2024)Gao, Xu, Ye, Liu, He, Fu, Mei, Wang, and Wu}]{gao2024designing}
Jiaxuan Gao, Shusheng Xu, Wenjie Ye, Weilin Liu, Chuyi He, Wei Fu, Zhiyu Mei, Guangju Wang, and Yi~Wu. 2024.
\newblock On designing effective rl reward at training time for llm reasoning.
\newblock \emph{arXiv preprint arXiv:2410.15115}.

\bibitem[{Guo et~al.(2025)Guo, Yang, Zhang, Song, Zhang, Xu, Zhu, Ma, Wang, Bi et~al.}]{deepseek-r1}
Daya Guo, Dejian Yang, Haowei Zhang, Junxiao Song, Ruoyu Zhang, Runxin Xu, Qihao Zhu, Shirong Ma, Peiyi Wang, Xiao Bi, et~al. 2025.
\newblock Deepseek-r1: Incentivizing reasoning capability in llms via reinforcement learning.
\newblock \emph{arXiv preprint arXiv:2501.12948}.

\bibitem[{He et~al.(2024)He, Luo, Bai, Hu, Thai, Shen, Hu, Han, Huang, Zhang et~al.}]{he2024olympiadbench}
Chaoqun He, Renjie Luo, Yuzhuo Bai, Shengding Hu, Zhen~Leng Thai, Junhao Shen, Jinyi Hu, Xu~Han, Yujie Huang, Yuxiang Zhang, et~al. 2024.
\newblock Olympiadbench: A challenging benchmark for promoting agi with olympiad-level bilingual multimodal scientific problems.
\newblock \emph{arXiv preprint arXiv:2402.14008}.

\bibitem[{Hendrycks et~al.(2021)Hendrycks, Burns, Kadavath, Arora, Basart, Tang, Song, and Steinhardt}]{hendrycks2021measuring}
Dan Hendrycks, Collin Burns, Saurav Kadavath, Akul Arora, Steven Basart, Eric Tang, Dawn Song, and Jacob Steinhardt. 2021.
\newblock Measuring mathematical problem solving with the math dataset.
\newblock \emph{arXiv preprint arXiv:2103.03874}.

\bibitem[{Kojima et~al.(2022)Kojima, Gu, Reid, Matsuo, and Iwasawa}]{kojima2022large}
Takeshi Kojima, Shixiang~Shane Gu, Machel Reid, Yutaka Matsuo, and Yusuke Iwasawa. 2022.
\newblock Large language models are zero-shot reasoners.
\newblock \emph{Advances in neural information processing systems}, 35:22199--22213.

\bibitem[{Lanham et~al.(2023)Lanham, Chen, Radhakrishnan, Steiner, Denison, Hernandez, Li, Durmus, Hubinger, Kernion et~al.}]{lanham2023measuring}
Tamera Lanham, Anna Chen, Ansh Radhakrishnan, Benoit Steiner, Carson Denison, Danny Hernandez, Dustin Li, Esin Durmus, Evan Hubinger, Jackson Kernion, et~al. 2023.
\newblock Measuring faithfulness in chain-of-thought reasoning.
\newblock \emph{arXiv preprint arXiv:2307.13702}.

\bibitem[{Lightman et~al.(2023)Lightman, Kosaraju, Burda, Edwards, Baker, Lee, Leike, Schulman, Sutskever, and Cobbe}]{PRM800k}
Hunter Lightman, Vineet Kosaraju, Yura Burda, Harri Edwards, Bowen Baker, Teddy Lee, Jan Leike, John Schulman, Ilya Sutskever, and Karl Cobbe. 2023.
\newblock Let's verify step by step.
\newblock \emph{arXiv preprint arXiv:2305.20050}.

\bibitem[{Liu et~al.(2024)Liu, Zhang, Luo, and Yao}]{liu2024augmenting}
Haoxiong Liu, Yifan Zhang, Yifan Luo, and Andrew Chi-Chih Yao. 2024.
\newblock Augmenting math word problems via iterative question composing.
\newblock \emph{arXiv preprint arXiv:2401.09003}.

\bibitem[{Liu et~al.(2025)Liu, Chen, Li, Pang, Du, and Lin}]{liu2025oatzero}
Zichen Liu, Changyu Chen, Wenjun Li, Tianyu Pang, Chao Du, and Min Lin. 2025.
\newblock There may not be aha moment in r1-zero-like training — a pilot study.
\newblock \url{https://oatllm.notion.site/oat-zero}.
\newblock Notion Blog.

\bibitem[{Luo et~al.(2024)Luo, Liu, Liu, Phatale, Lara, Li, Shu, Zhu, Meng, Sun et~al.}]{omega-prm}
Liangchen Luo, Yinxiao Liu, Rosanne Liu, Samrat Phatale, Harsh Lara, Yunxuan Li, Lei Shu, Yun Zhu, Lei Meng, Jiao Sun, et~al. 2024.
\newblock Improve mathematical reasoning in language models by automated process supervision.
\newblock \emph{arXiv preprint arXiv:2406.06592}.

\bibitem[{MAA(2024)}]{aime}
MAA. 2024.
\newblock \href {https://maa.org/math-competitions/american-invitational-mathematics-examination-aime} {American invitational mathematics examination - aime}.

\bibitem[{Min et~al.(2024)Min, Chen, Jiang, Chen, Deng, Hu, Tang, Wang, Cheng, Song et~al.}]{min2024imitate}
Yingqian Min, Zhipeng Chen, Jinhao Jiang, Jie Chen, Jia Deng, Yiwen Hu, Yiru Tang, Jiapeng Wang, Xiaoxue Cheng, Huatong Song, et~al. 2024.
\newblock Imitate, explore, and self-improve: A reproduction report on slow-thinking reasoning systems.
\newblock \emph{arXiv preprint arXiv:2412.09413}.

\bibitem[{Muennighoff et~al.(2025)Muennighoff, Yang, Shi, Li, Fei-Fei, Hajishirzi, Zettlemoyer, Liang, Cand{\`e}s, and Hashimoto}]{muennighoff2025s1}
Niklas Muennighoff, Zitong Yang, Weijia Shi, Xiang~Lisa Li, Li~Fei-Fei, Hannaneh Hajishirzi, Luke Zettlemoyer, Percy Liang, Emmanuel Cand{\`e}s, and Tatsunori Hashimoto. 2025.
\newblock s1: Simple test-time scaling.
\newblock \emph{arXiv preprint arXiv:2501.19393}.

\bibitem[{o1~Team(2024)}]{skyworkopeno12024}
Skywork o1~Team. 2024.
\newblock \href {https://huggingface.co/Skywork} {Skywork-o1 open series}.
\newblock \url{https://huggingface.co/Skywork}.

\bibitem[{OpenAI(2024)}]{openai2024learning}
OpenAI. 2024.
\newblock Learning to reason with llms.
\newblock \url{https://openai.com/index/learning-to-reason-with-llms/}.
\newblock Accessed: Month Day, Year.

\bibitem[{Park et~al.(2024)Park, Liu, Gong, and Choi}]{park2024ensembling}
Sungjin Park, Xiao Liu, Yeyun Gong, and Edward Choi. 2024.
\newblock Ensembling large language models with process reward-guided tree search for better complex reasoning.
\newblock \emph{arXiv preprint arXiv:2412.15797}.

\bibitem[{Setlur et~al.(2024)Setlur, Nagpal, Fisch, Geng, Eisenstein, Agarwal, Agarwal, Berant, and Kumar}]{setlur2024rewarding}
Amrith Setlur, Chirag Nagpal, Adam Fisch, Xinyang Geng, Jacob Eisenstein, Rishabh Agarwal, Alekh Agarwal, Jonathan Berant, and Aviral Kumar. 2024.
\newblock Rewarding progress: Scaling automated process verifiers for llm reasoning.
\newblock \emph{arXiv preprint arXiv:2410.08146}.

\bibitem[{Snell et~al.(2024)Snell, Lee, Xu, and Kumar}]{snell2024scaling}
Charlie Snell, Jaehoon Lee, Kelvin Xu, and Aviral Kumar. 2024.
\newblock Scaling llm test-time compute optimally can be more effective than scaling model parameters.
\newblock \emph{arXiv preprint arXiv:2408.03314}.

\bibitem[{Team et~al.(2024)Team, Georgiev, Lei, Burnell, Bai, Gulati, Tanzer, Vincent, Pan, Wang et~al.}]{team2024gemini}
Gemini Team, Petko Georgiev, Ving~Ian Lei, Ryan Burnell, Libin Bai, Anmol Gulati, Garrett Tanzer, Damien Vincent, Zhufeng Pan, Shibo Wang, et~al. 2024.
\newblock Gemini 1.5: Unlocking multimodal understanding across millions of tokens of context.
\newblock \emph{arXiv preprint arXiv:2403.05530}.

\bibitem[{Team(2025)}]{kimi_k1p5}
Kimi Team. 2025.
\newblock Kimi k1.5: Scaling reinforcement learning with llms.

\bibitem[{Team(2024{\natexlab{a}})}]{qwq}
Qwen Team. 2024{\natexlab{a}}.
\newblock \href {https://qwenlm.github.io/blog/qwq-32b-preview/} {Qwen team. qwq: Reflect deeply on the boundaries of the unknown, 2024b}.

\bibitem[{Team(2024{\natexlab{b}})}]{qwen2.5}
Qwen Team. 2024{\natexlab{b}}.
\newblock \href {https://qwenlm.github.io/blog/qwen2.5/} {Qwen2.5: A party of foundation models}.

\bibitem[{Touvron et~al.(2023)Touvron, Martin, Stone, Albert, Almahairi, Babaei, Bashlykov, Batra, Bhargava, Bhosale et~al.}]{touvron2023llama}
Hugo Touvron, Louis Martin, Kevin Stone, Peter Albert, Amjad Almahairi, Yasmine Babaei, Nikolay Bashlykov, Soumya Batra, Prajjwal Bhargava, Shruti Bhosale, et~al. 2023.
\newblock Llama 2: Open foundation and fine-tuned chat models.
\newblock \emph{arXiv preprint arXiv:2307.09288}.

\bibitem[{Uesato et~al.(2022)Uesato, Kushman, Kumar, Song, Siegel, Wang, Creswell, Irving, and Higgins}]{uesato2022solving}
Jonathan Uesato, Nate Kushman, Ramana Kumar, Francis Song, Noah Siegel, Lisa Wang, Antonia Creswell, Geoffrey Irving, and Irina Higgins. 2022.
\newblock Solving math word problems with process-and outcome-based feedback.
\newblock \emph{arXiv preprint arXiv:2211.14275}.

\bibitem[{Wang et~al.(2024{\natexlab{a}})Wang, Li, Shao, Xu, Dai, Li, Chen, Wu, and Sui}]{mathshepherd}
Peiyi Wang, Lei Li, Zhihong Shao, Runxin Xu, Damai Dai, Yifei Li, Deli Chen, Yu~Wu, and Zhifang Sui. 2024{\natexlab{a}}.
\newblock Math-shepherd: Verify and reinforce llms step-by-step without human annotations.
\newblock In \emph{Proceedings of the 62nd Annual Meeting of the Association for Computational Linguistics (Volume 1: Long Papers)}, pages 9426--9439.

\bibitem[{Wang et~al.(2024{\natexlab{b}})Wang, Li, Wu, Luo, Hou, Yu, and Shang}]{mips}
Zihan Wang, Yunxuan Li, Yuexin Wu, Liangchen Luo, Le~Hou, Hongkun Yu, and Jingbo Shang. 2024{\natexlab{b}}.
\newblock Multi-step problem solving through a verifier: An empirical analysis on model-induced process supervision.
\newblock \emph{arXiv preprint arXiv:2402.02658}.

\bibitem[{Wei et~al.(2022)Wei, Wang, Schuurmans, Bosma, Xia, Chi, Le, Zhou et~al.}]{wei2022chain}
Jason Wei, Xuezhi Wang, Dale Schuurmans, Maarten Bosma, Fei Xia, Ed~Chi, Quoc~V Le, Denny Zhou, et~al. 2022.
\newblock Chain-of-thought prompting elicits reasoning in large language models.
\newblock \emph{Advances in neural information processing systems}, 35:24824--24837.

\bibitem[{Zhang et~al.()Zhang, Zhoubian, Hu, Yue, Dong, and Tang}]{zhang2406rest}
Dan Zhang, Sining Zhoubian, Ziniu Hu, Yisong Yue, Yuxiao Dong, and Jie Tang.
\newblock Rest-mcts*: Llm self-training via process reward guided tree search, 2024a.
\newblock \emph{URL https://arxiv. org/abs/2406.03816}.

\bibitem[{Zhang et~al.(2025)Zhang, Zheng, Wu, Zhang, Lin, Yu, Liu, Zhou, and Lin}]{PRMlessons}
Zhenru Zhang, Chujie Zheng, Yangzhen Wu, Beichen Zhang, Runji Lin, Bowen Yu, Dayiheng Liu, Jingren Zhou, and Junyang Lin. 2025.
\newblock The lessons of developing process reward models in mathematical reasoning.
\newblock \emph{arXiv preprint arXiv:2501.07301}.

\bibitem[{Zhang et~al.(2022)Zhang, Zhang, Li, and Smola}]{zhang2022automatic}
Zhuosheng Zhang, Aston Zhang, Mu~Li, and Alex Smola. 2022.
\newblock Automatic chain of thought prompting in large language models.
\newblock \emph{arXiv preprint arXiv:2210.03493}.

\bibitem[{Zheng et~al.(2024)Zheng, Zhang, Zhang, Lin, Lu, Yu, Liu, Zhou, and Lin}]{zheng2024processbench}
Chujie Zheng, Zhenru Zhang, Beichen Zhang, Runji Lin, Keming Lu, Bowen Yu, Dayiheng Liu, Jingren Zhou, and Junyang Lin. 2024.
\newblock Processbench: Identifying process errors in mathematical reasoning.
\newblock \emph{arXiv preprint arXiv:2412.06559}.

\end{thebibliography}

\clearpage
\appendix
\section{Solution Reformatting}
\label{sec:solution_reformatting}
We use the method described in Section Section\ref{sec:division_solutions} to segment the open-sorce sample. An example of the solution reformation is shown in Figure \ref{fig:reformat_example_1}. After segmentation, sentences at the same step are shown in the same color. 

\begin{figure*}[ht]
\begin{tcolorbox}[colframe=black, colback=gray!10, coltitle=black, rounded corners, boxrule=0.3mm, width=\textwidth, halign=left]
\textbf{Problem:} The chord AB subtends an arc of the circle equal to $120^{circ}$ ). Point C lies on this arc, and point D lies on the chord AB. Additionally, AD = 2, BD = 1, and DC = $\sqrt{2}$. Find the area of triangle ABC.

\textbf{Solution after reformation:}
\textcolor{mydarkgreen}{
Alright, let's see here. I've got this geometry problem to solve. So, there's a circle with a chord AB that subtends an arc of 120 degrees. There's a point C on this arc, and a point D on chord AB such that AD is 2, BD is 1, and DC is sqrt(2). I need to find the area of triangle ABC.}
\vspace{15pt}
\textcolor{myorange}{First, I should probably draw a diagram to visualize this. So, I've got a circle, chord AB with points A and B on the circle, and the arc from A to B that's 120 degrees. Point C is somewhere on that arc, and point D is on AB such that AD is 2 and BD is 1, making AB a total of 3 units. DC is sqrt(2), which is interesting.}

\vspace{15pt} 

\textcolor{mydarkgreen}{Since AB subtends an arc of 120 degrees, that means the central angle corresponding to arc AB is 120 degrees. In a circle, the measure of an inscribed angle is half the measure of the central angle that subtends the same arc. So, angle ACB, which is inscribed and subtended by arc AB, should be 60 degrees.}

\vspace{15pt} 

\textcolor{mydarkgreen}{Wait, no. Actually, angle ACB is not necessarily 60 degrees. Let me think again. The central angle is 120 degrees, so the inscribed angle subtended by the same arc would be half of that, which is 60 degrees. But angle ACB is an inscribed angle subtended by arc AB, so yes, it should be 60 degrees.}

\vspace{15pt} 

\textcolor{myorange}{But hold on, point C is on the arc AB, so angle ACB is indeed 60 degrees. Now, I need to find the area of triangle ABC. To find the area, I can use the formula (1/2)*base*height, or (1/2)*ab*sin(C), where a and b are sides and C is the included angle.}

\vspace{15pt} 

\textcolor{myorange}{But I don't know the lengths of sides AC and BC yet. Maybe I can find those using the Law of Sines or Cosines.}

\vspace{15pt} 

\textcolor{mydarkgreen}{First, I need to find the lengths of AC and BC. Since I have a circle with chord AB of length 3, and central angle 120 degrees, I can find the radius of the circle.}

\vspace{15pt} 

\textcolor{mydarkgreen}{The formula for the length of a chord is 2*r*sin(theta/2), where theta is the central angle in radians. Wait, theta is in degrees here, so I need to be careful with units.}

\vspace{15pt} 

\textcolor{mydarkgreen}{Let me convert 120 degrees to radians. 120 degrees is (120/180)*pi = (2/3)pi radians.}

\vspace{15pt} 

\textcolor{mydarkgreen}{So, chord length AB = 2*r*sin((120)/2) = 2*r*sin(60 degrees). Sin 60 degrees is sqrt(3)/2, so AB = 2*r*(sqrt(3)/2) = r*sqrt(3).}

\vspace{15pt} 

\textcolor{mydarkgreen}{But AB is given as AD + DB = 2 + 1 = 3 units. So, r*sqrt(3) = 3, which means r = 3/sqrt(3) = sqrt(3).} \textcolor{mydarkgreen}{Okay, so the radius of the circle is sqrt(3).}
\end{tcolorbox}
\caption{An Example of solution reformation (part 1).}
\label{fig:reformat_example_1}
\end{figure*}

\begin{figure*}[!t]
\begin{tcolorbox}[colframe=black, colback=gray!10, coltitle=black, rounded corners, boxrule=0.3mm, width=\textwidth, halign=left]
\textcolor{myorange}{Now, I need to find AC and BC. Since C is on the arc AB that's 120 degrees, the angle at the center is 120 degrees, so the arc AC and arc CB must add up to 120 degrees. But I don't know their individual measures yet.}
\vspace{15pt} 
\textcolor{myorange}{Wait, point C is a specific point on the arc AB, but I don't have more information about its position. Maybe I need to use point D and the given lengths to find more information.}

\vspace{15pt} 

\textcolor{myorange}{Point D is on AB such that AD = 2 and DB = 1, and DC = sqrt(2). So, triangle DBC has sides DB = 1, DC = sqrt(2), and angle at D I don't know yet.}

\vspace{15pt} 

\textcolor{mydarkgreen}{Coordinates of A: (sqrt(3), theta)}

\vspace{15pt} 

\textcolor{mydarkgreen}{Coordinates of B: (-sqrt(3)/2, 3/2)}

\vspace{15pt} 

\textcolor{mydarkgreen}{Coordinates of C: (sqrt(2),1)}

\vspace{15pt} 

\textcolor{mydarkgreen}{Using shoelace formula again:}

\vspace{15pt} 

\textcolor{mydarkgreen}{Area = (1/2)|x1(y2 - y3) + x2(y3 - y1) + x3(y1 - y2)|}

\vspace{15pt} 

\textcolor{mydarkgreen}{Plug in:}

\vspace{15pt} 

\textcolor{mydarkgreen}{= (1/2)|sqrt(3)*(3/2 - 1) + (-sqrt(3)/2)*(1 - 0) + sqrt(2)*(0 - 3/2)|}

\vspace{15pt} 

\textcolor{mydarkgreen}{= (1/2)|sqrt(3)*(1/2) - sqrt(3)/2 - 3*sqrt(2)/2|}

\vspace{15pt} 

\textcolor{mydarkgreen}{= (1/2)| (sqrt(3)/2 - sqrt(3)/2) - 3*sqrt(2)/2 |}

\vspace{15pt} 

\textcolor{mydarkgreen}{= (1/2)| 0 - 3*sqrt(2)/2 |}

\vspace{15pt} 

\textcolor{mydarkgreen}{= (1/2)*(3*sqrt(2)/2)}

\vspace{15pt} 

\textcolor{mydarkgreen}{= (3*sqrt(2))/4}

\vspace{15pt} 

\textcolor{myorange}{Okay, so maybe it's correct. The area of triangle ABC is (3*sqrt(2))/4.}

\end{tcolorbox}
\caption{An Example of solution reformation (part 2).}
\label{fig:reformat_example_2}
\end{figure*}

\section{Prompt Template for LLM Annotation}
\label{sec:Prompt Template for LLM Judger}
The prompt template for LLM judger is shown in Figure \ref{fig:prompt_template_judger}.
\begin{figure*}[ht]
\begin{tcolorbox}[colframe=black, coltitle=black, rounded corners, boxrule=0.3mm, width=\textwidth, halign=left]
You are a mathematical expert. The user will provide a math problem, a step-by-step solution process, and the GT answer. First, you need to extract the short final answer from the GT Answer. Then, carefully check the user's step-by-step solution process and assign a score of either 0 or 1 for each step. 

You need to carefully examine the correctness of each step and provide a brief explanation. If there is an error in the current solution step, such as a calculation error, a derivation error, or a logical error, the score should be 0. If the current solution step is error-free, follow these rules to assign a score:

\begin{enumerate}
    \item \textbf{Error Propagation:} If there is an error in the preceding steps and the current step does not a new problem-solving idea or perform a proper correction, the score should be 0. For example, if STEP K contains an error, and STEP K+1 continues analyzing based on STEP K's error, then STEP K+1 is also considered incorrect, and the score should be 0.
    \item \textbf{Error Cessation:} If there is an error in the preceding steps, but the current step introduces a new, correct solution approach, the score should be 1. If the current step corrects the error from the previous steps, the score should also be 1. For example,
    \begin{itemize}
        \item STEP K contains an error.
        \item STEP K+1 continues the analysis based on the error.
        \item STEP K+2 corrects the previous error or introduces a problem-solving idea. 
    \end{itemize}
    In this case, STEP K and STEP K+1 should be scored as 0, and STEP K+2 should be scored as 1.
    \item Final Step: For the final step in the solution, if the answer obtained does not match the short answer, the score should be 0. If it matches the short answer, the score should be 1.
\end{enumerate}

Your response format should be in json format: \newline 
\phantom{AAAA}[\newline
\phantom{AAAAAAAA}\{ \newline
\phantom{AAAAAAAAAAAA}"STEP 0": 1(int), \newline
\phantom{AAAAAAAAAAAA}"Reason": xxxx(str) \newline
\phantom{AAAAAAAA}\}, \newline
\phantom{AAAAAAAA}\{ \newline
\phantom{AAAAAAAAAAAA} "STEP 1": 1(int), \newline
\phantom{AAAAAAAAAAAA} "Reason": xxxx(str)  \newline
\phantom{AAAAAAAA}\} \newline
\phantom{AAAAAAAA}... \newline
\phantom{AAAA}] \newline

\textbf{Note:} When analyzing the solution process, it is important to remain objective and rational, and not be misled by the description of the solution steps.

\end{tcolorbox}
\caption{The prompt template for LLM judger.}
\label{fig:prompt_template_judger}
\end{figure*}

\section{Training Details}
\subsection{SFT}
\label{Appendix: SFT ED}
The detailed training parameters for generator are provided in Table \ref{tab:sft_parameters}.
\begin{table}[ht]
\centering
\begin{tabular}{@{}l|c@{}}
\toprule
\textbf{Hyperparameter} & \textbf{Value}   \\
\hline
learning rate  & 1e-5   \\
epochs & 3 \\
batch size & 24  \\ 
max length & 16384 \\
\bottomrule
\end{tabular}
\caption{Hyperparameters of SFT}
\label{tab:sft_parameters}
\end{table}

\label{sec:experment_training_detail}
\subsection{PRM Training}
\label{Appendix: PRM ED}
The detailed training parameters in PRM training are provided in Table \ref{tab:PRM_training_parameters}.
\begin{table}[ht]
\centering
\begin{tabular}{@{}l|c@{}}
\toprule
\textbf{Hyperparameter} & \textbf{Value}   \\
\hline
learning rate  & 1e-6   \\
epochs & 1 \\
batch size & 256  \\ 
max length & 10240 \\
\bottomrule
\end{tabular}
\caption{Hyperparameters of PRM training}
\label{tab:PRM_training_parameters}
\end{table}
\section{Case Study:  An Annotation Result of LLM judger}
To better illustrate the PRM training data, an example case is presented in Figure \ref{fig:case-part1}. The model's reasoning process consists of 11 steps. The "content" field represents the model's reasoning at each step, "Score" indicates the evaluation by o1 for that step (1 for correct, 0 for incorrect), and "Reason" provides the rationale behind o1's scoring.
\label{sec:case}
\begin{figure*}[ht]
\begin{tcolorbox}[colframe=black, coltitle=black, rounded corners, boxrule=0.3mm, width=\textwidth, halign=left]

\textbf{Problem:}  A cross-country team completed a training run, and each student's performance was measured. Evelyn ran 4.5 distance units in 1.25 time units. Briana ran 4 distance units in 2.5 time units. Carla ran 5.2 distance units in 4.25 time units. Debra ran 2.8 distance units in 5.6 time units. Angela ran 1.4 distance units in 6.8 time units. Based on these results, which student had the second highest average speed?

\textbf{Solution:}
\begin{itemize}
    \item Step 1
    \begin{itemize}
        \item \textbf{Content:} Okay, so I've got this cross-country team's training run data. There are five students: Evelyn, Briana, Carla, Debra, and Angela. Each has recorded their distance and time. I need to figure out who has the second highest average speed.
        \item \textbf{Score:} 1.
        \item \textbf{Reason:} This step correctly introduces the problem, identifying the five students (Evelyn, Briana, Carla, Debra, and Angela) and the need to determine who has the second highest average speed. 
    \end{itemize}
    \item Step 2
    \begin{itemize}
        \item \textbf{Content:} First, I need to understand what average speed means. Average speed is the total distance traveled divided by the total time taken. So, I need to look at each student's data and see how far they went and how long it took them. The data gives me the distance traveled and time taken for each student.
        \item \textbf{Score:} 1.
        \item \textbf{Reason:} This step correctly explains that average speed is calculated by dividing the total distance traveled by the total time taken.
    \end{itemize}
    \item Step 3
    \begin{itemize}
        \item \textbf{Content:} Let's start with Evelyn. She traveled 4.5 distance units in 1.25 time units. Briana traveled 4 distance units in 2.5 time units. Carla traveled 5.2 distance units in 4.25 time units. Debra traveled 2.8 distance units in 5.6 time units. Angela traveled 1.4 distance units in 6.8 time.
        \item \textbf{Score:} 1. 
        \item \textbf{Reason:} Correctly lists all the data points given in the problem: Evelyn (4.5 distance units in 1.25 time units), Briana (4 distance units in 2.5 time units), Carla (5.2 distance units in 4.25 time units), Debra (2.8 distance units in 5.6 time units), and Angela (1.4 distance units in 6.8 time units).
    \end{itemize}
    \item Step 4
    \begin{itemize}
        \item \textbf{Content:} To find average speed, I need to calculate the total distance each ran and the total time it took, then divide distance by time. Based on the given data, it seems like each student maintained a constant speed since we only have their final distance and time measurements.
        \item \textbf{Score:} 1.
        \item \textbf{Reason:} This step correctly explains the approach to calculate average speed by dividing distance by time for each student. 
    \end{itemize}

\end{itemize}

\end{tcolorbox}
\caption{An example of an annotation result from the LLM judger (part 1).}
\label{fig:case-part1}
\end{figure*}

\begin{figure*}[ht]
\begin{tcolorbox}[colframe=black, coltitle=black, rounded corners, boxrule=0.3mm, width=\textwidth, halign=left]
\begin{itemize}
    \item Step 5
    \begin{itemize}
        \item \textbf{Content:} So, for Evelyn, from time 0 to 1.25 units, she went from 0 to 4.5 units, so her speed would be 4.5 divided by 1.25. Let me calculate that: 4.5 / 1.25 = 3.6 units per time unit. Similarly, Briana went from 0 to 4 units in 2.5 time units, so her speed is 4 / 2.5 = 0.625 units per time unit.
        \item \textbf{Score:} 0.
        \item \textbf{Reason:} This step incorrectly calculates Briana's speed, 4/2.5=1.6.
    \end{itemize}
    \item Step 6
    \begin{itemize}
        \item \textbf{Content:} Carla went from 0 to 5.2 units in 4.25 time units, so her speed is 5.2 / 4.25 = approximately 1.225 units per time unit. Debra went from 0 to 2.8 units in 5.6 time units, so her speed is 2.8 / 5.6 = 0.5 units per time unit. Angela went from 0 to 1.4 units in 6.8 time units, so her speed is 1.4 / 6.8 $\approx$ 0.205 units per time unit. So, comparing these speeds, Evelyn has the highest average speed at 3.6 units per time unit, second is Carla at 1.225, then Briana at approximately 0.625, Debra at 0.5, and finally Angela at approximately 0.205.
        \item \textbf{Score:} 0.
        \item \textbf{Reason:} Based on the incorrect calculation results in Step 5, the ranking is performed. This step is also considered incorrect.
    \end{itemize}
    \item Step 7
    \begin{itemize}
        \item \textbf{Content:} Wait a minute, but I should make sure I'm interpreting the data correctly. Maybe these numbers represent different aspects of the run. If these numbers represent the entire run, then the distance would be the total distance traveled, and the time would be the total time taken. Let's double-check: Evelyn is 3.6, Briana is 0.625, Carla is approximately 1.225, Debra is 0.5, Angela is 0.205. Since 3.6 >  1.225 > 0.625 > 0.5 > 0.205, then the second should be Carla.
        \item \textbf{Score:} 0. 
        \item \textbf{Reason:} Based on the incorrect calculation results from Step 5, the ranking is performed incorrectly.
    \end{itemize}
    \item Step 8
    \begin{itemize}
        \item \textbf{Content:} Wait, let’s recheck to prevent mistakes. For Evelyn: 4.5 / 1.25, or 4.5 / (5/4) = 4.5 * (4/5) = (4.5 * 4)/5 = 18/5 = 3.6, yes. For Briana: 2.5 / 4 = 25/40 = 0.625, yes. For Carla: 5.2 / 4.25 $\approx$ 1.2235, yes. For Debra: 2.8 / 5.6 = 0.5, yes. Angela: 1.4 / 6.8 $\approx$ 0.205, yes. The calculations are correct, and the order seems correct.
        \item \textbf{Score:} 0.
        \item \textbf{Reason:} When calculating Briana's average speed, the numerator and denominator are reversed.
    \end{itemize}

\end{itemize}

\end{tcolorbox}
\caption{An example of an annotation result from the LLM judger (part 2).}
\label{fig:case-part2}
\end{figure*}

\begin{figure*}[ht]
\begin{tcolorbox}[colframe=black, coltitle=black, rounded corners, boxrule=0.3mm, width=\textwidth, halign=left]
\begin{itemize}
    \item Step 9
    \begin{itemize}
        \item \textbf{Content:} Wait, wait, is the calculation really correct? Let’s analyze carefully. 4.5 / 1.25 = 3.6 — that’s okay. But 4 / 2.5 = 0.625? That’s wrong — 2.5 should not be greater than 4. 4 / 2.5 = 1.6 units per time unit, not 0.625 like I incorrectly calculated before!  So the correct ordering of speeds is: 3.6 (Evelyn) > 1.6 (Briana) > 1.224 (Carla) > 0.5 (Debra) > 0.206 (Angela). 
        \item \textbf{Score:} 1.
        \item \textbf{Reason:} Correctly recalculates and verifies all average speeds: Evelyn (3.6), Briana (1.6), Carla (1.225), Debra (0.5), and Angela (0.205). Properly rank them in descending order and correctly identify that Briana has the second highest average speed.
    \end{itemize}
    \item Step 10
    \begin{itemize}
        \item \textbf{Content:} To confirm, let's list the average speeds again: Evelyn: 4.5 / 1.25 = 3.6, Briana: 4 / 2.5 = 1.6, Carla: 5.2 / 4.25 $\approx$ 1.225, Debra: 2.8 / 5.6 = 0.5,  Angela: 1.4 / 6.8 $\approx$ 0.205, So, ordering them: 3.6 (Evelyn), 1.6 (Briana), 1.225 (Carla), 0.5 (Debra), 0.2059 (Angela). Therefore, the second highest is Briana with approximately 1.6. 
        \item \textbf{Score:} 1.
        \item \textbf{Reason:} Provides a final confirmation of the calculations and ranks, correctly stating that the second highest average speed belongs to Briana.
    \end{itemize}
    \item Step 11
    \begin{itemize}
        \item \textbf{Content:} Ha, I know the final answer, the student with the second highest average speed is Briana. Final answer: Briana.
        \item \textbf{Score:} 1.
        \item \textbf{Reason:} Correctly states the final answer as Briana.
    \end{itemize}

\end{itemize}

\end{tcolorbox}
\caption{An example of an annotation result from the LLM judger (part 3).}
\label{fig:case-part3}
\end{figure*}

\section{Human Annotation Details}
\label{sec:Human Annotation Details}
We employ individuals with bachelor's and graduate degrees to manually assess the annotation accuracy of different LLMs on the step-level test set. First, we provide training for the annotators to ensure they have a comprehensive understanding of the prompts used for LLM-based annotation in Appendix \ref{sec:Prompt Template for LLM Judger}. To facilitate a thorough comprehension of errors in the reasoning process, we classify mathematical reasoning errors into two categories: operation errors and conceptual errors. Operation errors encompass mistakes in formula calculations, derivations, and similar computational inaccuracies. In contrast, conceptual errors involve incorrect reasoning directions, such as misinterpreting the problem or improperly applying mathematical formulations.

For each entry in the step-level test set, annotators receive a file containing five fields: the question, the ground-truth answer, each step of the solution, the rationale provided by the LLM judge for each step, and the judgment result for each step. A step is considered correctly scored if both the rationale and the scoring outcome are deemed reasonable. Conversely, if either the rationale or the scoring outcome is found to be unreasonable, the scoring is marked as incorrect. The annotation accuracy of different LLMs is shown in Table \ref{tab:llm_annotation_acc}.

\begin{table}[ht]
\centering
\begin{tabular}{@{}lc@{}}
\toprule
\multicolumn{1}{c}{\textbf{Model}} & \textbf{Annotation Accuracy }     \\
\hline
gpt-4o-2024-08-06,  & 0.668\\
claude-3.5-sonnet-v2  & 0.726 \\
o1  & \textbf{0.963} \\
\bottomrule
\end{tabular}
\caption{The annotation accuracy of different models.}
\label{tab:llm_annotation_acc}
\end{table}

\section{Generalization of our PRM}
To further assess the generalization capability of our PRM, we conducted experiments on the AIME24 dataset under challenging conditions. Specifically, for each problem, we employ the DeepSeek-R1-Distill-Qwen-1.5B model to generate 32 candidate solutions. From these, we select 10 solutions per problem, ensuring that no more than 2 of the 10 were correct. This selection strategy increases the task difficulty for the PRM, as it must identify the correct solutions from a pool dominated by incorrect ones.

We evaluated the performance of various PRMs using two rule-based segmentation methods :
\begin{itemize}
    \item \textbf{Segmentation based on Reflection Words (SRW)}: This method segments the response content using reflection words and merges short segments to form more coherent units.
    \item \textbf{Segmentation based on Double Newlines (SDN)}: Double newline characters serve as delimiters to segment the response.
\end{itemize}
Experimental results are shown in Table \ref{tab:generation}. Notably, even when the generation model is not our SFT model and segmentation is performed using simple rule-based approaches, our PRM consistently achieves superior performance in terms of PRM@10 on the AIME24 dataset. These results demonstrate the robustness and strong generalization ability of our PRM across different segmentation strategies and generation models.
\begin{table}[ht]
\centering
\begin{tabular}{@{}lcc@{}}
\toprule
\multicolumn{1}{c}{\textbf{Model}} & \textbf{SRW} & \textbf{SDN}    \\
\hline
Qwen2.5-PRM-7B  & 0.367 & 0.333  \\
MathShepherd-7B  & 0.267 & 0.300 \\
Skywork-PRM-7B & 0.300 & 0.300 \\
Ours & \textbf{0.533}  & \textbf{0.500}  \\
\bottomrule
\end{tabular}
\caption{The PRM@10 of different PRMs under different segmentation methods.}
\label{tab:generation}
\end{table}

\section{Performance on Larger Models}
To strengthen our conclusions, we select Qwen2.5-14B-Base as the base model and apply our method to train a new PRM. As there are currently no open-source PRMs with comparable parameter sizes for direct comparison, we establish baselines by training Qwen2.5-14B-Base on the open-source PRM800K and MathShepherd datasets. The experimental results of various models on the PRM@64 metric are presented in Table \ref{tab:larger_model_size}. Our dataset achieves the best performance.

\begin{table}[ht]
\centering
\begin{tabular}{@{}lccc@{}}
\toprule
\textbf{Dataset} & \textbf{MATH} & \textbf{OBen} & \textbf{AIME24}    \\
\hline
PRM800K  & 0.810 & 0.502 & 0.233 \\
MathShepherd  & 0.770 & 0.455 & 0.200 \\
Ours & 0.964  & \textbf{0.536} & \textbf{0.333} \\
\bottomrule
\end{tabular}
\caption{The performance of different PRM trainset on Qwen2.5-14B-Base.}
\label{tab:larger_model_size}
\end{table}

\section{Robust Performance Scaling with Dataset Complexity}
\label{Appendix H}
While our PRM demonstrates modest improvements on simpler datasets, its true potential emerges when applied to more challenging problems that demand extensive reasoning and iterative refinement.

To quantify reasoning complexity across different datasets, we analyzed three key indicators: (1) average tokens, (2) average reasoning steps, and (3) average reflection tokens (RT). Reflection tokens encompass specific terms indicating reflection processes, such as "rethink" and "reconsider", with the vocabulary set derived from \cite{liu2025oatzero}. To measure our method's effectiveness, we calculated our PRM's relative advantage, defined as the difference between our PRM's PRM@64 and the mean PRM@64 of all baseline methods.

As illustrated in Table \ref{tab:performance_scaling}, there is a clear correlation between dataset complexity and our method's performance advantage. Specifically, as problems require longer reasoning chains and more frequent reflection steps, the performance gap between our PRM and baseline methods widens significantly. This scaling behavior provides strong evidence that our approach is particularly effective for complex reasoning tasks that demand extended CoT reasoning.

\begin{table}[ht]
\centering
\resizebox{\linewidth}{!}{
\begin{tabular}{@{}lcccc@{}}
\toprule
\textbf{} & \textbf{GSM8K} & \textbf{MATH500} & \textbf{OBen} & \textbf{AIME24}    \\
\hline
Difficulty Level  & Grade-school & High-school & Competition & Competition \\
Average Tokens  & 2339 & 5376 & 8150 & 12908 \\
Average Steps  & 6.9 & 13.1 & 17.8 & 25.8 \\
Average RT  & 2.5 & 5.7 & 8.3 & 12.0 \\
\hline
Advantage & 0.4\% & 6.0\% & 7.1\% & 8.0\%\\
\bottomrule
\end{tabular}}
\caption{Statistical metrics of reasoning processes across different datasets, and the advantage of our PRM compared to other PRMs on different datasets.}
\label{tab:performance_scaling}
\end{table}

\end{document}